\documentclass{article}
\usepackage{spconf,amsmath,graphicx}
\usepackage{url}            
\usepackage{booktabs}       
\usepackage{amsfonts}       
\usepackage{nicefrac}       
\usepackage{microtype}      
\usepackage{xcolor}         
\usepackage{amsmath}
\usepackage{subcaption}
\usepackage{caption}
\usepackage{graphicx}
\usepackage{cite}

\title{Sub-Graph Learning for Spatiotemporal Forecasting via Knowledge Distillation}
\name{Mehrtash Mehrabi and Yingxue Zhang}
\address{Huawei Noah’s Ark Lab}
\begin{document}

\maketitle

\begin{abstract}
    One of the challenges in studying the interactions in large graphs is to learn their diverse pattern and various interaction types.
    Hence, considering only one distribution and model to study all nodes and ignoring their diversity and local features in their neighborhoods, might severely affect the overall performance.
    Based on the structural information of the nodes in the graph and the interactions between them, the main graph can be divided into multiple sub-graphs.  
    This graph partitioning can tremendously affect the learning process, however the overall performance is highly dependent on the clustering method to avoid misleading the model.
    In this work, we present a new framework called KD-SGL to effectively learn the sub-graphs, where we define one global model to learn the overall structure of the graph and multiple local models for each sub-graph.
    %
    We assess the performance of the proposed framework and evaluate it on public datasets.
    Based on the achieved results, it can improve the performance of the state-of-the-arts spatiotemporal models with comparable results compared to ensemble of models with less complexity. 
\end{abstract}

\section{Introduction}


Recently there have been many studies on graph learning and specially spatiotemporal forecasting \cite{pal2021rnn, li2021spatial, song2020spatial, fang2021spatial}.
As the graph expands and includes more nodes, the interactions between nodes becomes more complicated and highly challenging to learn the spatiotemporal dependencies between all nodes.
Hence, the prediction model should be capable of precisely learning the various interaction types between nodes and find a policy to relate them to the underlying time series. 
However, all the existing studies consider only one single model to learn the node interactions across the graph where all the parameters are shared among all nodes equally which makes it unable to learn the node behaviour in local neighborhoods and distinguish different interaction types.


One simple solution to learn the node-specific features and the local interactions is partitioning the main large graph into several sub-graphs and study them independently.
However, the information loss caused by isolating the sub-graphs and ignoring the interactions between them can degrade the overall performance.
The learning procedure is highly dependent on the differences and relationships between the sub-graphs which should be considered during the learning process.
The main two questions in the problem of sub-graph learning, are 1) how to divide the main graph into sub-graphs and investigate them? and 2) how to use the interactions between the sub-graphs to enrich the model?






In this work, we investigate the problem of sub-graph learning which has not been explicitly studied in the literature.
We propose a new framework where the main large graph is partitioned into several sub-graphs based on their temporal and spatial proximity and learn an individual model for each graph separately. 
In order to improve the learning capability and incorporating the interactions between graphs, we introduce knowledge distillation (KD)-based sub-graph learning, called KD-SGL.
The KD-based approach can improve the learning capability by incorporating several sub-graph models and the interactions between them.

In the following sections, first in Sec.~\ref{sec:related}, we present some of related work.
We formulate the sub-graph learning problem in Sec.~\ref{sec:ProblemForm} and propose our model structure.
We evaluate KD-SGL in Sec.~\ref{sec:exp} by some of the existing datasets on traffic data.
Finally, we conclude the paper in Sec.~\ref{sec:conc}.

\section{Related Work}\label{sec:related}
In this section, first we present some of the recent works on spatiotemporal forecasting and then briefly summarize the recent studies on multi-task learning which is the closest topic in the literature to the sub-graph learning.

\subsection{Spatiotemporal Forecasting}
Our work is adapted for multivariate and spatiotemporal forecasting using deep learning and graph neural networks (GNNs).
Spatiotemporal forecasting has been extensively studied in the literature and neural network-based techniques have started to offer the best predictive performance for multivariate time-series prediction \cite{bao2017deep, qin2017dual, lai2018modeling, guo2018multi, chang2018memory, sen2019think, oreshkin2019n, smyl2020hybrid}.
Numerous algorithms have been proposed that combine GNNs with temporal neural network architectures \cite{yu2017spatio, huang2020lsgcn, zheng2020gman, park2020st, bai2020adaptive, oreshkin2021fc, li2017diffusion}. 

The authors in \cite{yu2017spatio} propose a model that instead of applying regular convolutional and recurrent units, formulates the problem on graphs and build the model with complete convolutional structures, which enable much faster training speed with fewer parameters.
In order to enhance the traditional graph convolutional network, the authors in \cite{bai2020adaptive} propose to use node adaptive parameter learning and data-adaptive graph generation modules for learning node-specific patterns and discovering spatial correlations from data, separately to capture node-specific spatial and temporal correlations in time-series data automatically without a pre-defined graph.
In \cite{oreshkin2021fc}, authors present a novel learning architecture called FC-GAGA, that can achieve a good performance without requiring knowledge of the graph.
It combines a fully connected time series model with temporal and graph gating mechanisms, that are both generally applicable and computationally efficient. 
In \cite{li2017diffusion}, the authors propose the diffusion convolutional recurrent neural network that captures the spatiotemporal dependencies by using bidirectional graph random walk and recurrent neural network.
They integrate the encoder-decoder architecture and the scheduled sampling technique to improve the long-term forecasting performance.

It should be noted that all of the above works consider only one common model between the nodes in the graph where all the nodes equally contribute to the model.
However, in sub-graph learning problem, our goal is to divide the main graph into several sub-graphs to better study the local interactions in each neighborhoods of the graph. 

\subsection{Multi-task Learning}
In conventional multi-task learning, the goal is to learn multiple task simultaneously, while in our problem the goal is to learn multiple sub-graphs of each single task.
Moreover, in multi-task learning models, the task boundaries are decided in advance, however in sub-graph learning the main graph should be split for training.
In the following, we present some of the recent works on multi-task learning.

The authors in \cite{fernando2017pathnet}, propose PathNet which is a neural network algorithm that uses embedded agents to discover which parts of the network to re-use for new tasks. 
Agents are pathways through the network which determine the subset of parameters that are used and updated by the forwards and backwards passes of the backpropogation algorithm. 
In \cite{mallya2018packnet}, authors have presented a method to pack multiple tasks into a single network with minimal loss of performance on prior tasks to obtain levels of accuracy comparable to the individually trained networks for each task. 
%
%
In \cite{shanahan2021encoders}, the authors have presented an architecture based on combining a pre-trained encoder, an ensemble of simple classifiers, and a particular activation / loss function pairing. 
It has shown that, these features independently helps to mitigate catastrophic forgetting in this setting, but they are most potent when working together. 

\section{KD-SGL Framework}\label{sec:ProblemForm}
Let us consider a graph where $N$ nodes are distributed across the graph and record the data of different time instances $t\in[1,T]$, where $T$ is the length of the window we investigate the graph.
At each time window, the data is measured by each node $n\in[1,N]$ and denoted as $\textbf{x}_n\in \mathbb{R}^T$ resulting in $\textbf{X}=[\textbf{x}_1, \cdots, \textbf{x}_N]^T\in \mathbb{R}^{N\times T}$ as the network data.
The final objective is to predict the next $H$ time instants status of nodes as $\tilde{\textbf{Y}} \in \mathbb{R}^{N\times H}$.

\begin{figure*}
    \vspace{-.2cm}
    \centering
    \includegraphics[width=.85\textwidth]{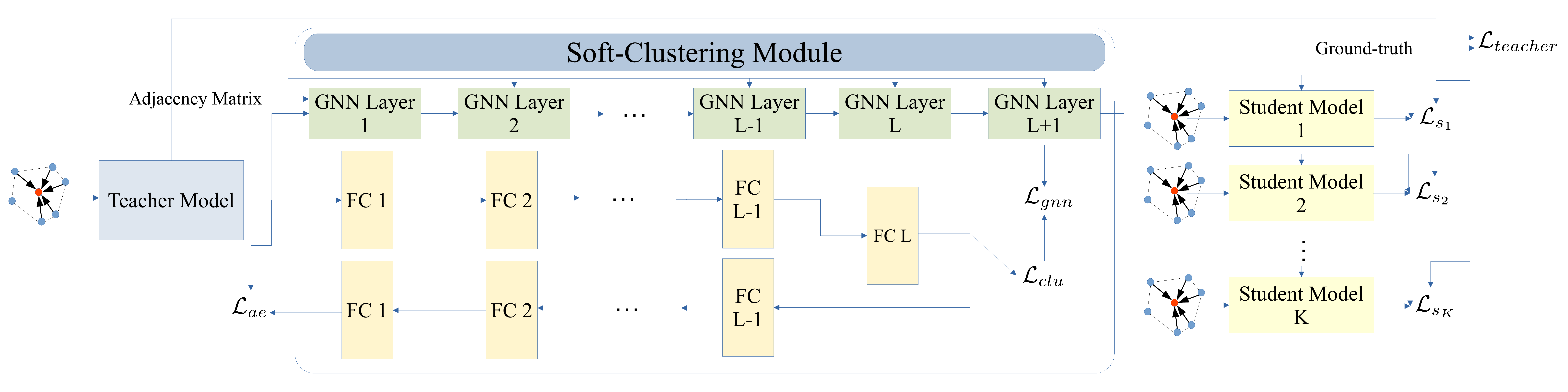}
    \vspace{-.4cm}
    \caption{End-to-End KD-based model structure}
    \label{fig:modelStructure}
    \vspace{-.6cm}
\end{figure*}

The properties of the nodes across the graph can be investigated both locally within local neighborhoods and globally by considering all the node interaction types in the graph.
As the graph expands, it becomes more crucial to consider local neighborhoods since the nodes become farther away from each other and include more interaction types.
KD-SGL can enhance the overall performance by investigating the node interactions both locally and globally through introducing global and local models, called teacher and student.
The overall structure of KD-SGL is illustrated in Fig.\ref{fig:modelStructure}, consisting of three main modules, including, a teacher model, node clustering, and several student models.
In the following, we present each component of KD-SGL in detail.




\subsection{Teacher Model}
Given a large graph formed by several nodes with highly diverse features and interaction types, KD-SGL first utilize all the nodes across the graph equally to train a global model called teacher.
The teacher model learns the nodes embedding matrix $\textbf{E}\in \mathbb{R}^{N\times d}$, where $d$ is the dimension of the embedding and make a prediction for the next $i$-th state of node $n$ as $\hat{y}_{n,T+i}$, where $i\in[1,H]$. 
Moreover, the teacher model is later used to supervise the local models (students) for each sub-graph.
In order to train the teacher model we minimize the following loss term defined as
\begin{equation}
    \mathcal{L}_{teacher} = \frac{1}{HN}\sum_{j=1}^{H}\sum_{i=1}^{N} |y_{i,T+i} - \hat{y}_{i,T+i}|.
\end{equation}

\subsection{Node-Clustering}
The goal of node-clustering is to define the sub-graphs and determine the contribution of nodes to each of them.
The sub-graphs are denied based on the spatial and temporal information of nodes investigated through the learnt nodes embedding by the teacher model and the adjacency matrix of the main graph.
Inspired by the clustering method introduced in \cite{bo2020structural}, we introduce our node-clustering approach in KD-SGL, where instead of hard-clustering, and fully isolating the sub-graphs, we perform soft-clustering, and determine how much each node should contribute to each sub-graph.
The node-clustering module is the integration of an autoencoder (AE) and GNN layers.
The AE is fed by the learned nodes embedding matrix $\textbf{E}$ and the output of its layer $l\ge 1$ is
\begin{equation}
    \textbf{A}^{(l)} = \phi(\textbf{U}^{(l)}\textbf{A}^{(l-1)} + \textbf{b}^{(l)}),
\end{equation}
where $\textbf{U}^{(l)}$ and $\textbf{b}^{(l)}$ are the weights and biases of layer $l$, respectively and $\textbf{A}^{(0)}=\textbf{E}$.
By choosing the proper weight matrices, the output of the encoder is generated as $\textbf{A}^{(L)}\in \mathbb{R}^{N\times K}$ to match the number of $K$ sub-graphs. 
The output of the decoder part is the reconstruction of the raw data $\hat{\textbf{E}}$ leading to the following loss function \vspace{-.2cm}
\begin{equation}
    \mathcal{L}_{ae} = \frac{1}{N}\sum_{i=1}^{N} ||\textbf{e}_i - \hat{\textbf{e}}_i||^2_2.
\end{equation}

For each layer of the encoder, there is a corresponding GNN layer fed by the encoded data of last layer, i.e. $\textbf{A}^{(l-1)}$, and the adjacency matrix $\textbf{W}$, as
\begin{equation}
    \textbf{Z}^{(l)} = \phi(\tilde{\textbf{D}}^{-\frac{1}{2}}\tilde{\textbf{W}}\tilde{\textbf{D}}^{-\frac{1}{2}} \tilde{\textbf{Z}}^{(l-1)}\textbf{U}^{(l-1)}),
\end{equation}
where $\tilde{\textbf{W}} = \textbf{W} + I$, $\tilde{\textbf{D}}$ is diagonal matrix and $\tilde{\textbf{D}}_{ii}=\sum_j\tilde{\textbf{W}}_{ij}$, and $\textbf{U}^{(l-1)}$ is the weight matrix of layer $l-1$.
Function $\phi(\cdot)$ is a non-linear function such as $\tanh$ and $\tilde{\textbf{Z}}^{(l-1)}$ is generated as 
\begin{equation}
    \tilde{\textbf{Z}}^{(l-1)} = (1-\psi) \textbf{Z}^{(l-1)} + \psi \textbf{A}^{(l-1)},
\end{equation}
where $\psi$ is the balance coefficient and here is set to 0.5.
It should be noted that $\textbf{Z}^{(0)} = \textbf{E}$ and hence $\tilde{\textbf{Z}}^{(0)} = \textbf{E}$.
The last GNN layer is the input of a multiple classification layer with a softmax function to map $N$ nodes to $K$ sub-graphs as
\begin{equation}
    \textbf{Z} = softmax(\tilde{\textbf{D}}^{-\frac{1}{2}}\tilde{\textbf{W}}\tilde{\textbf{D}}^{-\frac{1}{2}} \textbf{Z}^{(L)}\textbf{U}^{(L)}) \in \mathbb{R}^{N\times K}.
\end{equation}

We use the Student’s t-distribution as a kernel to measure the similarity between the data representation $\textbf{a}_i$ and the cluster center vector $\mu_j$ as follows
\begin{equation}
    q_{ij} = \frac{(1 + ||\textbf{a}_i - \mu_j||^2/v)^{-\frac{v+1}{v}}}{\sum_{j'}(1+||\textbf{a}_i - \mu_{j'}||^2/v)^{-\frac{v+1}{v}}},
\end{equation}
where $\textbf{a}_i$ is the $i$-th row of $\textbf{A}^{(L)}\in \mathbb{R}^{N\times K}$, $v$ is the degree of freedom of the student's t-distribution, and $\mu_j$ is initialized by K-means on the representations learned by pre-train AE. 
In order to make data representation closer to cluster centers and improving the cluster cohesion we calculate a target distribution $\textbf{P}$ as \vspace{-.2cm}
\begin{equation} 
    p_{ij} = \frac{q_{ij}^2/f_j}{\sum_{j'}q^2_{ij'}/f_{j'}},
\end{equation}
where $f_j=\sum_i q_{ij}$ are soft cluster frequencies. 
Then the clustering loss is defined as 
\begin{equation}
    \mathcal{L}_{clu} = KL(\textbf{P}||\textbf{Q}) = \sum_i \sum_j p_{ij} log\frac{p_{ij}}{q_{ij}}.
\end{equation}
The GNN also provide a clustering assignment distribution $\textbf{Z}$ (output of the classification layer) and we can use distribution $\textbf{P}$ to supervise it by defining the following loss function
\begin{equation}
    \mathcal{L}_{gnn} = KL(\textbf{P}||\textbf{Z}) = \sum_i \sum_j p_{ij}log\frac{p_{ij}}{z_{ij}}.
\end{equation}
\vspace{-1cm}
\subsection{Student Models}
In order to investigate the generated sub-graphs, we define local models, called students trained for each sub-graph based on the learnt weights $\textbf{Z}$.
The student models are supervised by the teacher model by a pre-defined imitation factor that determines how much student models should follow the teacher model.
Using the teacher model and the learnt distribution $\textbf{Z}$, $K$ student models are trained to make the final prediction for $N$ nodes. 
The teacher supervision through the KD structure can also help to learn less complex student models compared to the teacher model and benefit from the features already learnt by the complex teacher.
Hence, we can use a simple multiple layer perception (MLP) for student models with less parameters than the teacher model.
Each student model $\textbf{S}_k$, where $k\in[1,K]$ is then trained as
\begin{align}\nonumber
    \mathcal{L}_{s_k} =& (1-\rho) \Big(\frac{1}{HN}\sum_{i=1}^{H}\sum_{j=1}^{N} z_{jk}|y_{j,T+i} - {s_k}_{j,T+i}|\Big) + \\ & \rho \Big(\frac{1}{HN}\sum_{i=1}^{H}\sum_{j=1}^{N} z_{jk}|\hat{y}_{j,T+i} - {s_k}_{j,T+i}|\Big), 
\end{align}
where $\rho$ is the pre-defined KD imitation factor and $z_{ik}$ is the probability of node $i$ assigned to cluster $k$. 

\begin{figure*}[t!]
    \vspace{-.2cm}
    \centering
    \begin{subfigure}[t]{0.45\textwidth}
        \centering
        \includegraphics[height=1.4in]{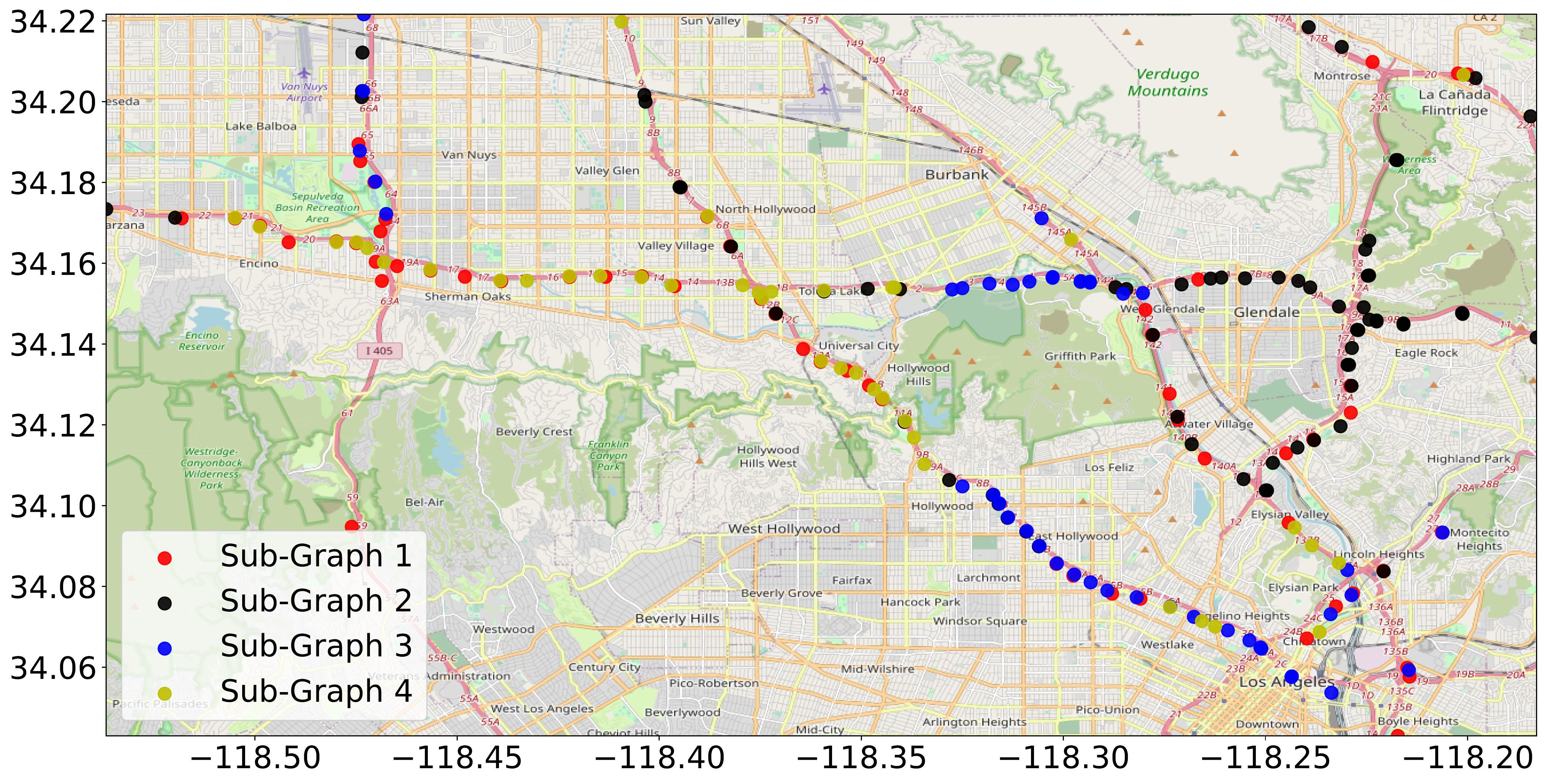}
        \vspace{-.2cm}
        \caption{Location of nodes from different sub-graphs}
        \vspace{-.2cm}
    \end{subfigure}%
    ~ 
    \begin{subfigure}[t]{0.45\textwidth}
        \centering
        \includegraphics[height=1.4in]{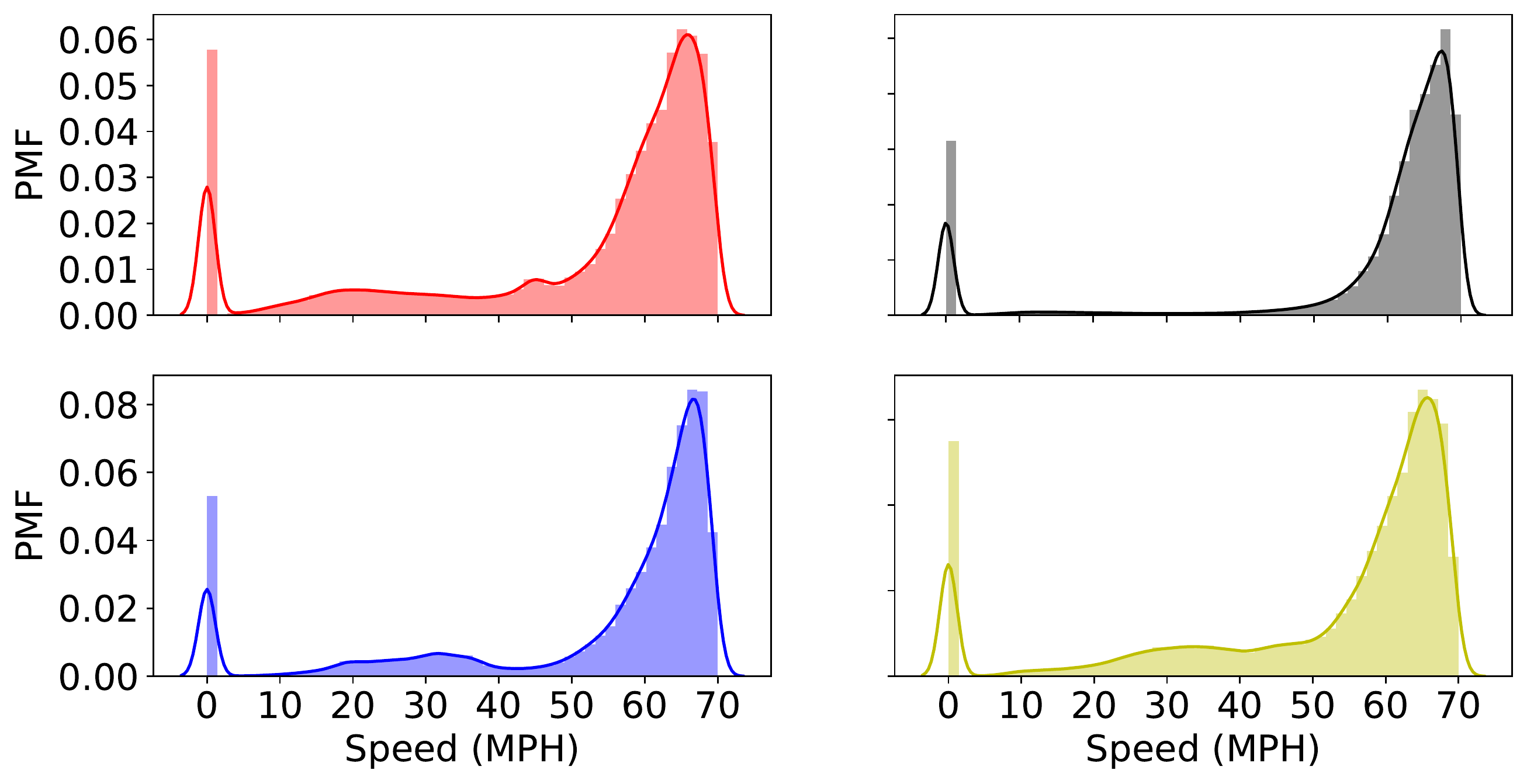}
        \vspace{-.2cm}
        \caption{Distribution of the traffic pattern in different sub-graphs}
        \vspace{-.2cm}
    \end{subfigure}
    ~ 
    \caption{Location and distribution of nodes from different sub-graphs generated by KD-SGL (FC-GAGA) for METR-LA dataset.}
    \vspace{-.5cm}
    \label{fig:node_clustering}
\end{figure*}

Considering all the loss sources, the final loss function of KD-SGL is computed as
\begin{align}
    \mathcal{L}_{tot} = \mathcal{L}_{teacher} + 
    & \mathcal{L}_{ae} + \alpha\mathcal{L}_{clu} + \beta\mathcal{L}_{gnn} +
     \sum_{k=1}^K \mathcal{L}_{s_k},
\end{align}
where $\alpha>0$ is a hyperparameter that balances the clustering optimization and local structure preservation of raw data and $\beta>0$ is a coefficient that controls the disturbance of GNN layers to the embedding space.
The final prediction is 
\begin{equation}
    \tilde{y}_{i,t} = \frac{1}{2}\Big(\hat{y}_{i,t} + \sum_{k=1}^K z_{i,k}\cdot y_{i,t}^{(k)}\Big),
\end{equation}
where $\hat{y}_{i,t}$ and $y_{i,t}^{(k)}$ are the predicted values for node $i$ at time $t$, by the teacher model and the $k$-th student, respectively.


\section{Experimental Results}\label{sec:exp}
We assess KD-SGL using two spatiotemporal models including FC-GAGA and DCRNN. 
We use two datasets with different size to study the impact of the size of the graph on the performance and sub-graph learning.
In the following first, we present the experimental setup and then evaluate KD-SGL in terms of the achieved performance and its complexity.

\vspace{-.2cm}
\subsection{Simulation Setup}
\textbf{Datasets}. We evaluate KD-SGL on two widely explored traffic datasets with different size including METR-LA and PEMS07 with $N=207$ and $N=883$ nodes, respectively. 
METR-LA consists of the traffic speed readings collected from loop detectors and aggregated over 5 minute intervals. 
It contains 34,272 time steps of 207 sensors collected in Los Angeles County over 4 months.
PEMS07 is collected from the Caltrans Performance Measurement System (PeMS) and aggregated into 5-minutes windows, resulting in 288 points in the traffic flow for one day.
It contains 28224 time steps of 883 sensors constructing a graph with 883 nodes and 866 edges.

\textbf{Training Process}. 
In order to train KD-SGL, the datasets are split into 70\% training, 10\% validation, and the final 20\% as the test sets.
For the following tests, for case of FC-GAGA, we use it with 4 layers as the teacher model in our KD-SGL framework.
We set the number of sub-graphs, i.e. $K$, to 4 and for each student model we use the verification set to find the best imitation factor $\rho$ and then test the performance by the test set.
The values of $\alpha$ and $\beta$ are set to 0.1. 

\textbf{Hardware Setup}. The following experiments are conducted on a system with 16 Gb of GPU and Cuda version 11.1 and tensorflow 2.2.

\vspace{-.2cm}
\subsection{Overall Performance and Complexity Analysis}

In Fig.~\ref{fig:node_clustering}, we show the locations of the nodes of different sub-graphs generated by the node clustering method for METR-LA dataset, where for each node the sub-graph with the highest score is selected. 
We also compare the distribution of nodes in different learnt sub-graphs, and as seen the distributions are close to each other but the details are different which are interpreted as global and local information.

Table \ref{tab:model_comparison} shows the comparison between the achieved results by KD-SGL framework using FC-GAGA and DCRNN as the teacher model by running several trials and compare 3 horizons $t=3, 6, 9$ and use three metrics to compare the results.
We compare the achieved performance results with the original results of the base models (FC-GAGA and DCRNN) and also their ensembles.
As seen, for the larger dataset we can observe a larger performance gap which shows the importance of sub-graph learning in KD-SGL.
We also compare the complexity of different models during the inference in Table~\ref{tab:complexity} which verifies the effectiveness of KD-SGL over ensemble approach.

\begin{table}[t!]
    \centering
    \resizebox{\columnwidth}{!}{
    \begin{tabular}{*{2}c | *{9}c}
        \toprule
        \multicolumn{2}{c}{} & \multicolumn{3}{c}{15 min} &  \multicolumn{3}{c}{30 min} & \multicolumn{3}{c}{60 min} \\ 
        Dataset & Model  & MAE & MAPE & RMSE & MAE & MAPE & RMSE & MAE & MAPE & RMSE\\ \hline
        {} & FC-GAGA & 2.75 & 7.23 & 5.35 & 3.11 & 8.59 & 6.35 & 3.52 & 10.16 & 7.33\\
        METR-LA & Ensemble (4 FC-GAGA) & \textbf{2.71} & \textbf{7.12} & \textbf{5.19} & \textbf{3.06} & \textbf{8.46} & \textbf{6.10} & \textbf{3.45} & \textbf{10.01} & \textbf{7.05}\\
        {} & KD-SGL (FC-GAGA)  & 2.73 & 7.20 & 5.26 & 3.07 & 8.49 & 6.17 & \textbf{3.45} & 10.05 & 7.11\\
        {} & DCRNN  & 2.77 & 7.30 & 5.38 & 3.15 & 8.80 & 6.45 & 3.60 & 10.50 & 7.59\\
        METR-LA & Ensemble (4 DCRNN) & 2.74 & 7.20 & 5.25 & 3.10 & 8.57 & 6.33 & 3.50 & 10.27 & 7.38\\
        {} & KD-SGL (DCRNN) & 2.75 & 7.24 & 5.29 & 3.12 & 8.61 & 6.35 & 3.52 & 10.29 & 7.41\\
        \hline \hline
        {} & FC-GAGA  & 19.63 & 8.29 & 31.89 & 21.42 & 9.07 & 35.17 & 23.50 & 10.03 & 38.73\\
        PEMS07 & Ensemble (4 FC-GAGA) & 19.37 & \textbf{8.15} & 31.25 & \textbf{20.65} & \textbf{8.70} & \textbf{33.70} & \textbf{22.69} & \textbf{9.68} & \textbf{37.07}\\
        {} & KD-SGL (FC-GAGA)  & \textbf{19.34} & 8.19 & \textbf{31.24} & 20.74 & 8.78 & 33.86 & 22.70 & 9.71 & 37.18 \\
        {} & DCRNN  & 19.69 & 8.34 & 31.93 & 21.47 & 9.10 & 35.19 & 23.50 & 10.12 & 38.79\\
        PEMS07 & Ensemble (4 DCRNN) & 19.41 & 8.24 & 31.36 & 20.93 & 8.86 & 34.64 & 22.91 & 9.81 & 38.07 \\
        {} & KD-SGL (DCRNN)  & 19.43 & 8.25 & 31.27 & 20.96 & 8.90 & 34.70 & 22.91 & 9.82 & 38.18 \\
        \bottomrule
    \end{tabular}}
    \vspace{-.3cm}
    \caption{Performance comparison of proposed KD-SGL framework with FC-GAGA and DCRNN models as the base.}
    \vspace{-.2cm}
    \label{tab:model_comparison}
\end{table}

\begin{table}[t!]
    \centering
    \resizebox{\columnwidth}{!}{
    \begin{tabular}{c|c|c|c}
    \hline
         Dataset & Model & Prediction Time (s) & Number of Parameters\\ \hline
         & FC-GAGA & 75.13 & 4.23M\\
         METR-LA & Ensemble (4 FC-GAGA) & 292.32 & 16.92M\\
         & KD-SGL(FC-GAGA) & 264.83 & 13.21M\\
         & DCRNN & 7.75 & .37M\\
         METR-LA & Ensemble (4 DCRNN) & 29.34 & 1.48M\\
         & KD-SGL(DCRNN) & 22.37 & 1.12M\\
         \hline
    \end{tabular}}
    \vspace{-.3cm}
    \caption{Complexity Analysis}
    \vspace{-.5cm}
    \label{tab:complexity}
\end{table}

\section{Conclusion}\label{sec:conc}
In this work, we propose KD-SGL, a novel framework for spatiotemporal forecasting that learns the spatial and temporal dependencies across the graph by partitioning it into several sub-graphs.
It introduce two types of models called teacher and student to learn the global and local structural information in the graph.
Extensive experiments on comprehensive traffic datasets demonstrates significant improvement of KD-SGL over existing models that study the graphs as a single graph.  
Our proposed model is generic and flexible, and any existing spatiotemporal forecasting model can be used as the teacher model to enhance the overall performance.
For future work, we plan to study the impact of loss terms on the final result and further optimize the clustering method.

\bibliographystyle{IEEEtran}
\bibliography{main.bbl}



\end{document}